\pdfoutput=1

\documentclass[11pt]{article}

\usepackage[]{acl}

\usepackage{times}
\usepackage{latexsym}
\usepackage{arydshln}
\usepackage[T1]{fontenc}

\usepackage[utf8]{inputenc}
\usepackage{booktabs}
\usepackage{array}
\usepackage{microtype}
\usepackage{multirow}
\usepackage{graphicx}
\usepackage{bbding}
\usepackage{amsmath}
\usepackage{enumitem}
\usepackage{enumerate}
\usepackage{subcaption}
\usepackage{colortbl}
\usepackage{float}
\title{Not All Contexts Are Equal: Teaching LLMs Credibility-aware Generation}
\author{Ruotong Pan${}^{1,2}$,
        Boxi Cao${}^{1,2,}$\thanks{~ Corresponding authors.},
        Hongyu Lin${}^{1}$,
        Xianpei Han${}^{1}$,\\
        {\bf Jia Zheng${}^{1,}$\footnotemark[1],}
        {\bf Sirui Wang${}^{3}$,}
        {\bf Xunliang Cai${}^{3}$,}
        {\bf Le Sun${}^{1}$} \\
          {${}^{1}$}Chinese Information Processing Laboratory, Institute of Software, \\ Chinese Academy of Sciences, Beijing, China \\
   ${}^{2}$University of Chinese Academy of Sciences, Beijing, China \\
  ${}^{3}$Meituan \\
  {\tt{\small \{panruotong2021, boxi2020, hongyu, xianpei, zhengjia\}@iscas.ac.cn}}\\
  {\tt{\small \{sunle\}@iscas.ac.cn}}
{\tt{\small \{wangsirui,caixunliang\}@meituan.com}}
}
\definecolor{low}{RGB}{68, 114, 196}
\definecolor{mid}{RGB}{124, 124, 124}
\definecolor{high}{RGB}{216, 119, 154}
\definecolor{olivegreen}{RGB}{107, 142, 35}
\definecolor{graybg}{gray}{0.85}

\begin{document}
\maketitle

\begin{abstract}

The rapid development of large language models has led to the widespread adoption of Retrieval-Augmented Generation (RAG), which integrates external knowledge to alleviate knowledge bottlenecks and mitigate hallucinations.
However, the existing RAG paradigm inevitably suffers from the impact of \emph{flawed information} introduced during the retrieval phrase, thereby diminishing the reliability and correctness of the generated outcomes.
In this paper, we propose Credibility-aware Generation (CAG), a universally applicable framework designed to mitigate the impact of flawed information in RAG.
At its core, CAG aims to equip models with the ability to discern and process information based on its credibility.
To this end, we propose an innovative data transformation framework that generates data based on credibility, thereby effectively endowing models with the capability of CAG.
Furthermore, to accurately evaluate the models' capabilities of CAG, we construct a comprehensive benchmark covering three critical real-world scenarios.
Experimental results demonstrate that our model can effectively understand and employ credibility for generation, significantly outperform other models with retrieval augmentation, and exhibit robustness despite the increasing noise in the context.
\footnote{
Our code, benchmark, and models are available at \url{https://github.com/panruotong/CAG}}
\end{abstract}

\section{Introduction}
In recent years, Large Language Models (LLMs) \citep{brown2020language,openai2023gpt4, touvron2023llama,anil2023palm} have experienced significant growth and demonstrated excellent performance in multiple domains \citep{kojima2022large,thirunavukarasu2023large,ziems2023can,min2023recent}.
With the ascendancy of LLMs, Retrieval-Augmented Generation (RAG) has attracted significant interest. RAG mitigates the knowledge bottleneck of LLMs by incorporating externally retrieved documents into their generation process. This inclusion helps diminish the occurrences of hallucinations and misinformation during generation, thereby substantially enhancing the quality of output from LLMs \citep{petroni-etal-2021-kilt, zhu2021retrieving, mallen-etal-2023-trust}.


\begin{figure}
    \centering
    \setlength{\belowcaptionskip}{-0.3333cm}
    \includegraphics[scale=0.48]{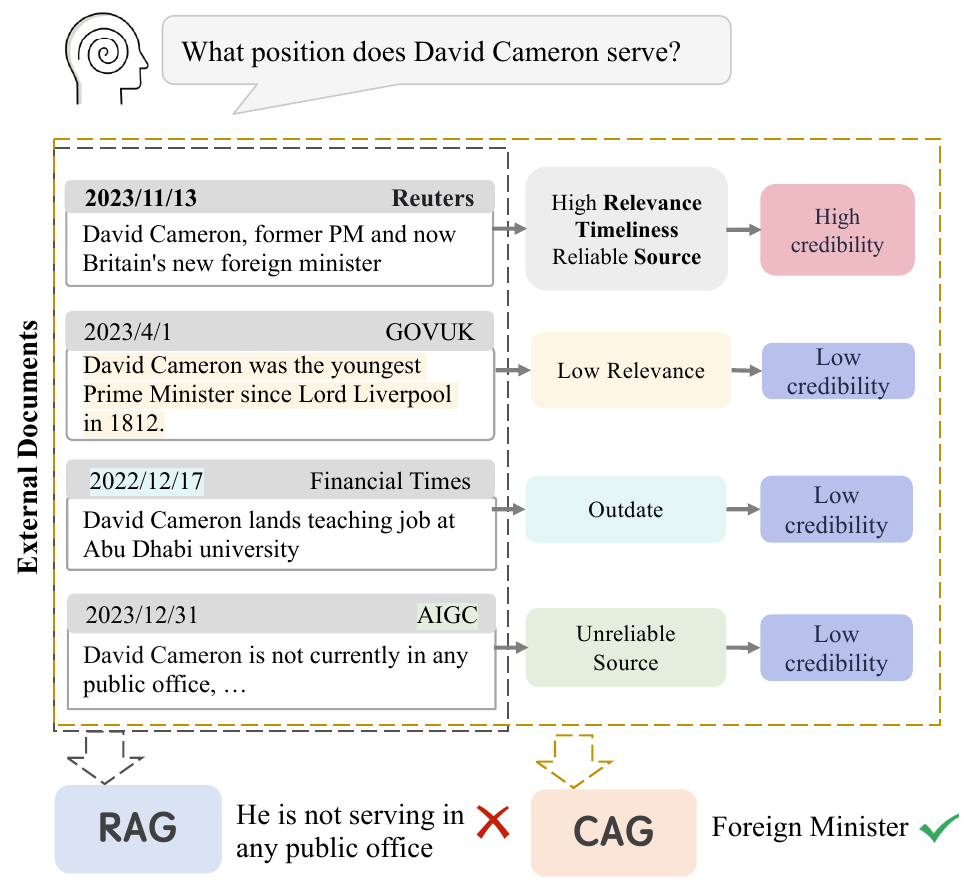}
    \caption{The comparison between Retrieval-Augmented Generation (RAG) and Credibility-aware Generation (CAG). Incorporating credibility into the model aids in mitigating errors caused by \emph{flawed information} introduced from the retrieval process.}
\end{figure}


However, RAG for large language models remains significantly impacted by flawed information.
This is mainly because the retrieval process often provides noisy, outdated, and incorrect contexts which adversely affects RAG, substantially reducing its effectiveness.
Specifically, previous research~\cite{shi2023large, chen_benchmarking_2023} has found that LLMs are highly sensitive to noise, which impacts LLMs' capacity to discern and trust accurate information, ultimately affecting the outcomes they generate.
Furthermore, due to the temporal insensitivity of LLMs ~\citep{su2022improving, zhao2024set}, these models struggle to discern outdated information solely based on their internal knowledge. More critically, because LLMs are trained on extensive collections of historical text, there's an inherent risk that outdated information will align with the models' internal knowledge bases. This alignment can encourage LLMs to favor and perpetuate outdated information.
Besides, the prevalence of misinformation on the current web poses a significant challenge for large models, which struggle to identify misinformation using only their inherent knowledge \citep{xie_adaptive_2023, pan2023risk}. This difficulty makes them susceptible to misinformation, leading to the generation of incorrect answers. 
Therefore, flawed information, characterized by noisy, outdated, and incorrect information, has substantial negative effects on RAG.
From a cognition perspective, a common approach humans adopt to combat flawed information is to assess the credibility of external information~\cite{burgoon_interactivity_2000}. For humans, credibility refers to the acceptability based on the quality of the information, its source, and subjective evaluation. However, LLMs relying solely on internal knowledge to assess information credibility are unstable and unreliable \citep{xie_adaptive_2023}. Therefore, we aim to guide LLMs' acceptance of information by utilizing external indicators of credibility.
We introduce \textbf{Credibility-aware Generation} (\textbf{CAG}), a universally applicable framework designed to address flawed information encountered during RAG. 
At its core, CAG seeks to equip models with the ability to discern and process information based on  credibility. 
By assigning different credibility to information based on its relevance, timeliness, and the reliability of its source, and explicitly distinguishing them in the input, CAG significantly mitigates the issues arising from flawed information.

Unfortunately, we have discovered that existing LLMs are not inherently sensitive to directly provided credibility in the prompt. This deficiency restricts their capacity to optimally employ credibility for discerning and processing information.
To endow models with the capability of CAG, we propose a novel data transformation framework. This framework transforms existing Question Answering (QA) datasets into data that integrates credibility, which can be employed to guide the model for credibility-based generation.
Specifically, our process comprises two core steps: 
1) 
Multi-granularity credibility annotation, which assigns credibility to text units at both document and sentence levels by dividing retrieved documents into varying granularities.
2) Credibility-guided explanation generation, which prompts LLMs to generate credibility-guided explanations given questions, retrieved documents with credibility annotation and \emph{golden answers}.
Finally, we employ instruction fine-tuning to train the model, enabling it to generate responses based on credibility.

To rigorously assess the ability of the model's credibility-aware generation in managing flawed information, we construct a comprehensive benchmark encompassing various real-world scenarios, including open-domain QA, time-sensitive QA, and misinformation polluted QA.
In this benchmark, retrieval relevance, timeliness, and source authority are regarded as established measures of credibility.
Experimental results on multiple datasets across multiple scenarios demonstrate the efficacy of our approach in utilizing credibility. 
Our model significantly outperforms various prevalent RAG approaches applied to both open and closed-source LLMs of diverse scales. 
Additionally, it exhibits robust resilience against noisy documents, maintaining high performance even as alternative strategies suffer sharp declines.
All these results verify the effectiveness of the proposed CAG framework and corresponding training algorithm.

\begin{figure*}[th]
    \centering
    \setlength{\belowcaptionskip}{-0.5cm}
    \includegraphics[scale=0.53]{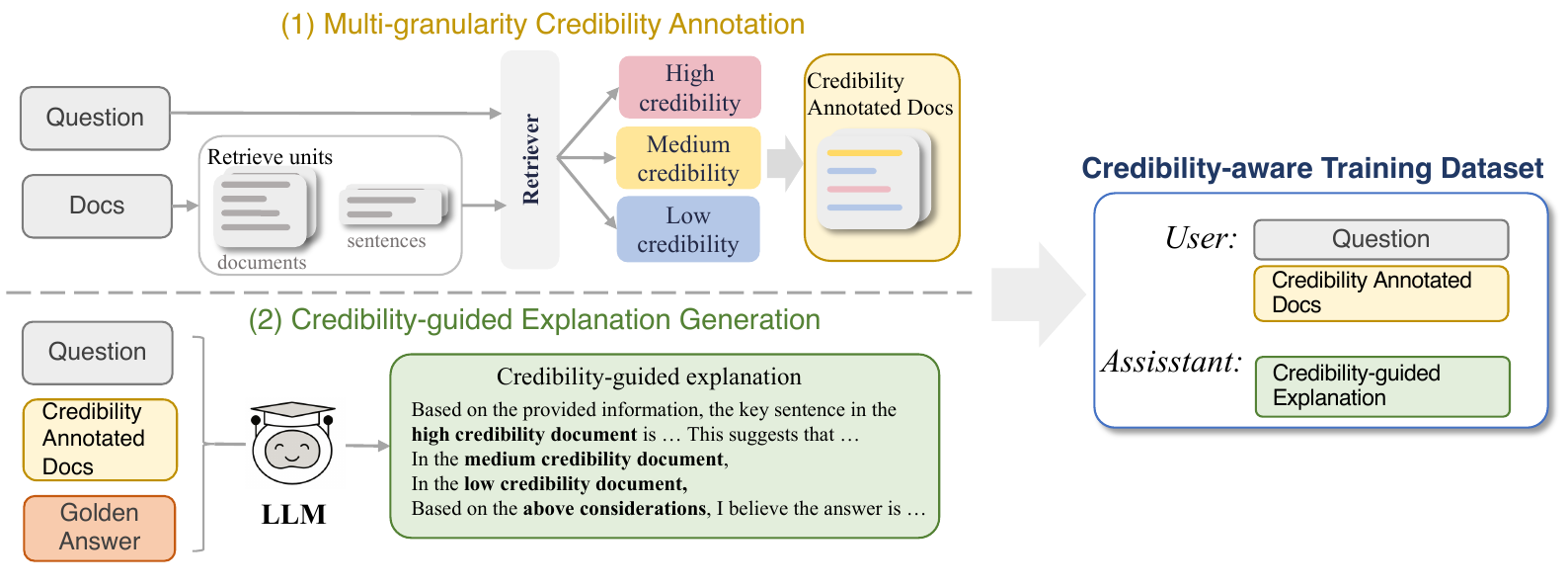}
    \caption{Overview of data transformation framework.  The training data is constructed by assigning credibility to contexts via multi-granularity credibility annotation (\S\ref{annotation}) and prompting LLM to produce credibility-guided explanations (\S\ref{explan}). The processed data is used to instruction fine-tuning (\S\ref{training}) to endow the model with the ability for Credibility-aware Generation.}
    \label{fig:enter-label}
\end{figure*}

The main contributions of this study are summarized as follows \footnote{We uploaded the code and datasets as supplemental mate-rials, which will be openly released after accepting.}:
\begin{itemize}
    \setlength{\itemsep}{2pt}
    \setlength{\parsep}{2pt}
    \setlength{\parskip}{1.8pt}
    \setlength{\topsep}{1pt}
    \item We present Credibility-aware Generation, a universal framework to handle the flawed information challenge in RAG.
    
    \item We propose a novel data transformation framework that transforms existing datasets into data annotated with credibility and guides models to generate responses based on credibility, thereby equipping the model with Credibility-aware Generation capability.
    
    \item We construct a comprehensive benchmark and evaluate model performance in credibility-aware generation, encompassing real-world scenarios of open-domain QA, time-sensitive QA, and misinformation polluted QA.

    \item Experimental evidences demonstrate that our model effectively understands and employs credibility to generate responses, significantly surpasses other RAG-based strategies, and maintains robustness despite the increasing noise in the context.
\end{itemize}


\section{Credibility-aware Generation}
Credibility-aware Generation is designed to enable models to discern and process information based on its credibility. 
Subsequently, we will provide formal definitions for both RAG and CAG, illustrating their divergence.
\label{define}
\paragraph{Definition}
In the Retrieval-Augmented Generation process, user input \(\boldsymbol{x}\) initiates the retrieval of a set of related documents \(D_{\boldsymbol{x}}\) from a large corpus \(C\) based on how closely these documents match the input. Then, it combines the input \(\boldsymbol{x}\) with these documents \(D_{\boldsymbol{x}}\) to generate responses \(\boldsymbol{y}\), formalized as \(\boldsymbol{y} = \mathrm{LM}([\boldsymbol{x}, D_{\boldsymbol{x}}])\), where \(\left[.,.\right]\) denotes the concatenation operation.

Compared to RAG, the Credibility-aware  Generation offers additional credibility for each document.
Initially, through credibility assessment based on various scenarios, each retrieved document has been assigned a level of credibility.
Then, these documents \(D_{\boldsymbol{x}}\) with their credibility  \(C\) are synthesized with the user input \(\boldsymbol{x}\) as augmented input. LM generates responses $\boldsymbol{y}$ based on this augmented input,
formally represented as $
    \boldsymbol{y} = \mathrm{LM} \left( \left[\boldsymbol{x}, \left\{ [c_i, d_i] \right\}_{i=1}^{|D_{\boldsymbol{x}}|} \right]\right)
$. 
This approach ensures that the generated responses not only incorporate the content of the documents but also consider the credibility of each document, thereby enhancing the reliability of responses.

\section{Teaching Model to Credibility-aware Generation}
\label{method}


In this section, we endow LLMs with the capability of CAG. A potential approach involves directly providing the credibility annotations of each document in the prompt. Unfortunately, as indicated in Table \ref{tab:results}, our experiments reveal that even advanced LLMs, such as ChatGPT, exhibit limited sensitivity to credibility. 
To this end, we introduce a novel data transformation framework. Through multi-granularity credibility annotation and credibility-guided explanation generation, we transform existing QA datasets into data that includes credibility annotations which can guide the model to generate credibility-based responses. Then, through instruction fine-tuning, we train the model to generate responses grounded in credibility assessments.


\subsection{Multi-granularity Credibility Annotation}\label{annotation}
To cater to the varied requirements for credibility across different scenarios and enhance the model's comprehension of credibility,  we collect training data including open-domain QA, machine reading comprehension, and dialogue datasets and propose a multi-granularity credibility annotation method.

First, we divide the retrieved documents to create a multi-granularity corpus, encompassing sentence and document levels. Then, the retriever assesses the match between each retrieval unit and the query, assigning a relevance score, and classifies documents into three levels: high, medium, and low, using either equal count or equal interval methods.
This approach of using levels instead of scores aims to simplify representation, thereby improving the model's understanding and providing a certain degree of fault tolerance.
Ultimately, we collect about 15k training data samples, all of which include documents with credibility annotations. The detailed composition of the training data is shown in the Appendix \ref{train_data}.




\subsection{Credibility-guided Explanation Generation}
\label{explan}
To facilitate the model's comprehension and effective utilization of credibility, we employ LLMs to generate credibility-guided explanations for the answers.

Given the limitations of current LLMs in comprehending credibility effectively, we design chain-of-thought prompts to guide LLMs to generate credibility-guided explanations given questions, retrieved documents with credibility  and \textbf{golden answers}. In this case, LLMs only need to generate coherent explanations based on the document containing the answers, without distinguishing between documents with different credibility to generate answers.
The credibility-guided explanation obtained includes an analysis that integrates both the credibility and the content of the documents, rather than merely focusing on deriving the answer.

Considering the accessibility and advanced capabilities, we employ GPT-3.5 for the generation of explanations.
In this way, we obtain high-quality answer explanations.
Then, we replace the original answers in the training data with credibility-guided explanations to form a novel QA dataset.
In this dataset, the inputs include questions and external documents annotated with credibility, while the outputs are credibility-guided explanations.

\subsection{Instruction Fine-tuning}
\label{training}

Through the two steps above, the training dataset obtained contains credibility, which can be used to facilitate arbitrary language models in gaining the capacity for CAG.
We fine-tune the language model on this dataset to empower the model to discern and process information according to its credibility. 
As defined by \citet{iyer2023optiml}, the loss function is as follows:
\begin{align*}
\small
\mathcal{L}(D_{\boldsymbol{x}}; \theta) = - \sum_{i=1}^{N} \log p_\theta\left(\boldsymbol{y}_i \mid \left[\boldsymbol{x}, \left\{[c_i, d_i]\right\}_{i=1}^{|D_{\boldsymbol{x}}|}\right], \boldsymbol{y}_{<i}\right)\end{align*}
\section{Credibility-aware Generation Benchmark}
\label{benchmark}

To rigorously evaluate the ability of credibility-aware model generation to handle flawed information, we construct the Credibility-aware Generation Benchmark (CAGB). This benchmark encompasses the following three specific scenarios where the integration of credibility is essential:

$\bullet$ \textbf{Open-domain QA} 
     aims to accurately answer questions on a wide variety of topics without being limited to any particular area. It encompasses a broad spectrum of real-world applications that urgently require the integration of external knowledge to enhance the LLMs' ability to address queries. This scenario necessitates the ability to effectively identify and process noise information.
     
$\bullet$ \textbf{Time-sensitive QA}
    aims to give accurate and current answers. It poses a challenge for LLMs due to the dynamic internet information. The inevitable inclusion of outdated documents when incorporating external sources further complicates matters. Even with timestamps provided for documents, LLMs may erroneously prioritize outdated documents. This situation underscores the critical need for credibility in time-sensitive QA.
    
$\bullet$ \textbf{Misinformation polluted QA} 
    aims to tackle the issue of ensuring accurate answers in an environment polluted with  misinformation. It presents a substantial challenge to LLMs, attributed to the misuse of LLMs and the consequent proliferation of fake news and misinformation  \citep{zhuo2023exploring, pan2023risk}. Consequently, it is crucial to take into account the quality and credibility of any introduced external information.
In the following, we will provide a detailed description of data construction for each scenario, and the statistics of CAGB are shown in the Table \ref{tab:statis}.
\subsection{Credibility Assessment}
We aim to establish a flexible credibility assessment mechanism that can be conveniently extended to consider additional factors and a broader range of application fields. In this benchmark, the credibility of the documents is evaluated by considering retrieval relevance, timeliness, and source reliability.
Specifically, we establish a foundation based on retrieval relevance, then make adjustments according to timeliness, and finally integrate the reliability of the source to determine credibility.
First, the retriever assigns relevance scores to documents based on query similarity. These relevance scores, which are distributed at equal intervals, enable to classify documents into three levels: high, medium, and low, collectively denoted as $R$.
Subsequently, the temporal difference $T$ between the query time and document publication is calculated, downgrading $R$ if $T$ surpasses a threshold. The formula integrating relevance and timeliness is as follows:

\scalebox{0.92}{$rt\_score(R,T) = \text{max}(R-\text{floor}(T/\text{threshold}),1)$}

Following this, the source reliability, denoted $S$, is customized to specific scenarios, similarly divided into three levels. 
Finally, we combine these factors, adopting the lower level as the credibility and the formula is expressed as follows:
\begin{align*}
    Cred=\mathrm{min}(rt\_score(R,T),S)
\end{align*}
    
In this way, the document of high credibility are concurrently characterized by high relevance, timeliness and source reliability. More details about the assessment can be seen in the Appendix \ref{cred_asses}.
\subsection{Open-domain QA}
Our research utilizes data from several challenging QA datasets with noisy documents. HotpotQA \citep{yang-etal-2018-hotpotqa} and 2WikiMHQA \citep{ho2020constructing} both require reasoning across multiple documents, and feature a high proportion of distracting documents.  Importantly, the data we utilize from HotpotQA is extracted from the dev subset, whereas our training dataset is derived from the train subset.
Musique \citep{Trivedi2021MM} questions are of higher complexity, with up to 90\% of distracting passages. ASQA \citep{stelmakh-etal-2022-asqa} is a long format QA dataset focused on ambiguous questions. RGB \citep{chen_benchmarking_2023} is a specialized benchmark used for evaluating the capabilities of models in the RAG scenario, with noise robustness being one of its aspects. We assign credibility to the documents provided in the dataset in terms of retrieval relevance. 

\begin{table}[tp]
\centering
\setlength{\belowcaptionskip}{-0.5cm}
\resizebox{.41 \textwidth}{!}{
\begin{tabular}{lccc}
\toprule
\textbf{Dataset}        & \textbf{\#Samples} & \textbf{\#Documents} & \multicolumn{1}{l}{\textbf{Noise Ratio}} \\ \midrule
\rowcolor{graybg}\multicolumn{4}{c}{\textit{Open-domain QA}}   \\
HotpotQA & 500       & 5000        & 0.8   \\
2WikiMHQA      & 500       & 5000        & 0.6-0.8    \\
MuSiQue        & 500       & 10000       & 0.9        \\
ASQA& 948       & 4740        & - \\
RGB & 300 & 11641 & 0.2-0.8\\ \midrule
\rowcolor{graybg}\multicolumn{4}{c}{\textit{Time-sensitive QA}}      \\
EvolvTempQA & 321       &  2247& 0.4-0.8\\ \midrule
\rowcolor{graybg}\multicolumn{4}{c}{\textit{Misinformation polluted QA}}        \\
NewsPollutedQA & 480 & 2400 & 0.5-0.75\\
\bottomrule
\end{tabular}
}
\caption{Statistics of CAGB, which includes 7 dataset derived from 3 scenarios.}
\label{tab:statis}
\end{table}
\subsection{Time-sensitive QA}
\label{time}
In order to construct a diverse, high-quality, and up-to-date news dataset, we annotate 321 time-sensitive questions along with their corresponding dates.
These questions originate from real-world scenarios, including news QA data from RealTime QA \citep{kasai2022realtime}, TAQA \citep{zhao2024set}, and questions adapted from news reports.
To simulate the simultaneous occurrence of varied information on the Internet, we use Google search API to retrieve each query, selecting 3 relevant documents and 4 distracting documents. The distracting documents are either irrelevant to the query or outdated. This approach to document selection is crafted to emulate the intricate and heterogeneous nature of real-world information landscapes. Each news includes its publication date, thereby aiding in the evaluation of its timeliness. 
For document credibility annotation, we assess credibility based on relevance and time gap between the document's publication and the posed question.
We ensure the accuracy of the answers by manually annotating. 

The obtained time-sensitive dataset with outdated document settings and credibility annotation is named EvolvingTempQA.

\begin{table*}[t]
\setlength{\belowcaptionskip}{-0.5cm}
\centering
\resizebox{0.95\textwidth}{!}{
\begin{tabular}{@{}lccccccc@{}}
\toprule
\multirow{2}{*}{\textbf{Model}}        & \multicolumn{5}{c}{\textbf{Open-domain QA}}     & \textbf{Time-sensitive QA} & \textbf{Misinfo polluted QA} \\ \cmidrule(l){2-8} & HotpotQA& 2WikiMHQA & MuSiQue & ASQA    & RGB  & EvolvingTempQA    &  NewsPollutedQA  \\ \midrule
\rowcolor{graybg} \multicolumn{8}{c}{\textit{retrieval-based}}\\ \midrule
ChatGPT          & 0.334    & 0.368   & 0.194   & 0.404       &    0.773   & 0.579  & 0.231    \\
LLaMA-2-7B       & 0.280   & 0.312   & 0.160   & 0.268       &   0.753   & 0.433  & 0.179    \\
Vicuna-7B        & 0.278   & 0.296   & 0.116   & 0.358       &   0.677   & 0.567  & 0.229    \\
Mistral-7B-Instruct & 0.288 & 0.270 & 0.106 & 0.300 & 0.713 & 0.598 & 0.204\\
LLaMA-2-13B      & 0.366   & 0.370    & 0.164   & 0.321       &   0.820   & 0.495  & 0.204    \\
LLaMA-2-70B      & 0.418   & 0.390    & 0.256   & 0.316       &   0.823   & 0.526  & 0.430    \\
vanilla IFT      & 0.324   & 0.245   & 0.270   & 0.157       &   0.650   & 0.592  & 0.329    \\ \midrule
\rowcolor{graybg} \multicolumn{8}{c}{\textit{ retrieval and reranking}}\\ \midrule
ChatGPT          & 0.396   & 0.394   & 0.216   & 0.388   & 0.790     & 0.632  & 0.427    \\
LLaMA-2-7B       & 0.302   & 0.376   & 0.200   & 0.375   & 0.730     & 0.526  & 0.265      \\
Vicuna-7B        & 0.355   & 0.306   & 0.164   & 0.494   & 0.757     & 0.620  & 0.275    \\
Mistral-7B-Instruct & 0.338 & 0.334 & 0.166 & 0.414 & 0.790 & 0.741 & 0.373\\
LLaMA-2-13B      & 0.370    & 0.372   & 0.180   & 0.390    & 0.823     & 0.561  & 0.308    \\
LLaMA-2-70B      & 0.422    & 0.504   & 0.320   & 0.388   & 0.833     & 0.570  & 0.306    \\
vanilla IFT      & 0.348   & 0.448   & 0.276   & 0.304   & 0.663     & 0.720  & 0.344    \\ \midrule
\rowcolor{graybg} \multicolumn{8}{c}{\textit{ retrieval and credibility}}      \\ \midrule
ChatGPT          & 0.422   & 0.402   & 0.182   & 0.440   & 0.807     & 0.673  & 0.408      \\
LLaMA-2-7B       & 0.376   & 0.176   & 0.140   & 0.394   & 0.713     & 0.486  & 0.213     \\
Vicuna-7B        & 0.349   & 0.266   & 0.091   & 0.490   & 0.740     & 0.642  & 0.279    \\
Mistral-7B-Instruct & 0.274 & 0.268 & 0.102 & 0.463 & 0.797 & 0.679 & 0.315 \\
LLaMA-2-13B      & 0.360    & 0.384   & 0.164   & 0.385   & 0.803     & 0.520  & 0.227    \\
LLaMA-2-70B      & 0.398   & 0.402   & 0.262   & 0.492   & 0.817 & 0.536  & 0.279    \\
vanilla IFT      & 0.372   & 0.334   & 0.204   & 0.305   & 0.663& 0.589& 0.383    \\ \hline

CAG-7B (\textit{ours})& \underline{0.509} & \underline{0.578} & {0.340} & {0.496}& 0.897 & 0.826 & 0.442 \\ 
CAG-13B (\textit{ours})& \textbf{0.514}& \textbf{0.604}& \textbf{0.408} & \textbf{0.525} & \textbf{0.917} & \underline{0.829} & \underline{0.483}\\
CAG-mistral-7B (\textit{ours})& 0.502 & 0.540 & \underline{0.384} & \underline{0.505} & \underline{0.900} & \textbf{0.835} & \textbf{0.613} \\
\bottomrule
\end{tabular}
}
\caption{Model performance in our CAGB benchmark.
The best/second best scores in each dataset are \textbf{bolded}/\underline{underlined}.
Our models substantially outperform previous strategies across all 3 scenarios in CAGB.
The results shown for EvolvingTempQA and RGB are at noise\_ratio setting of 0.8, while NewsPollutedQA is at noise\_ratio setting of 0.75. The results of other metrics on the ASQA dataset are shown in the Appendix \ref{appendix:asqa}.}
\label{tab:results}
\end{table*}

\subsection{Misinformation Polluted QA}\label{misinfo}
We create a up-to-date multiple-choice quiz dataset, comprising both real and fake news for each question.
The dataset construction bases on RealTime QA, utilizing weekly news quizzes from CNN and other news platforms. To maintain the dataset's real-time relevance, we select news from July 1, 2023, onwards, comprising 480 questions with four options and one supporting news item each. 

To simulate the generation of fake news, we first generated a claim using LLMs, based on a question and a randomly selected incorrect option.
This process transforms the question and incorrect option into a deceptive statement. Subsequently, we choose GPT-3.5 and Qwen \citep{bai2023qwen} as the generators for fake news, guiding them to generate texts of varying styles based on the claim, including news style and Twitter style. The prompts used and examples are detailed in the Appendix \ref{prompt}.
The fictitious news articles produced by LLMs, due to their authenticity being deliberately compromised, are classified as having low source reliability. Conversely, news articles from reputable news websites are considered to possess high source reliability. We set the ratio of fake news at 0.5, 0.67, and 0.75 to evaluate the robustness of model against  misinformation under various levels of pollution.

By simulating fake news generation, we create a  misinformation polluted QA dataset in the news domain, named NewsPollutedQA.


\section{Experiments} \label{exper}

To demonstrate the effectiveness of our framework in handling flawed information in real-world QA scenarios, we conduct comprehensive experiments within the CAGB. 
All these results verify the effectiveness of the CAG framework and the corresponding training algorithm. Additionally, our models maintain robustness even with the increasing noise in the context. 
In the following sections, we will discuss our experiments and conclusions in detail.

\subsection{Setup}

\paragraph{Baselines}
We compare our method with the following three strategies incorporated with 7 LLMs across various scales:
\begin{itemize}
    \setlength{\itemsep}{2pt}
    \setlength{\parsep}{2pt}
    \setlength{\parskip}{1.8pt}
    \item \textbf{Retrieval-based} concatenates documents from the dataset with questions as input.
    \item \textbf{Retrieval and reranking} employs an advanced reranking mechanism to reorder retrieved documents, giving priority to those with greater relevance \citep{xie_adaptive_2023}.
    \item  \textbf{Retrieval and credibility}  incorporates credibility as a prefix to the retrieved documents in the prompt, aiming to assess the model's ability to understand and utilize credibility.
\end{itemize}

We evaluate advanced models, including ChatGPT 
 \texttt{(gpt-3.5-turbo-0613)}, LLaMA-2-7B, 13B, 70B, Vicuna-7B-v1.5 and Mistral-7B-Instruct \citep{jiang2023mistral}. Additionally, we create a dataset mirroring the model training data but without credibility annotations and with initial answers, on which we fine-tune the LLaMA-2-7B model, and named the trained model vanilla IFT.

\begin{figure*}[t]
    \centering
    \setlength{\belowcaptionskip}{-0.4cm}
    \includegraphics[scale=0.35]{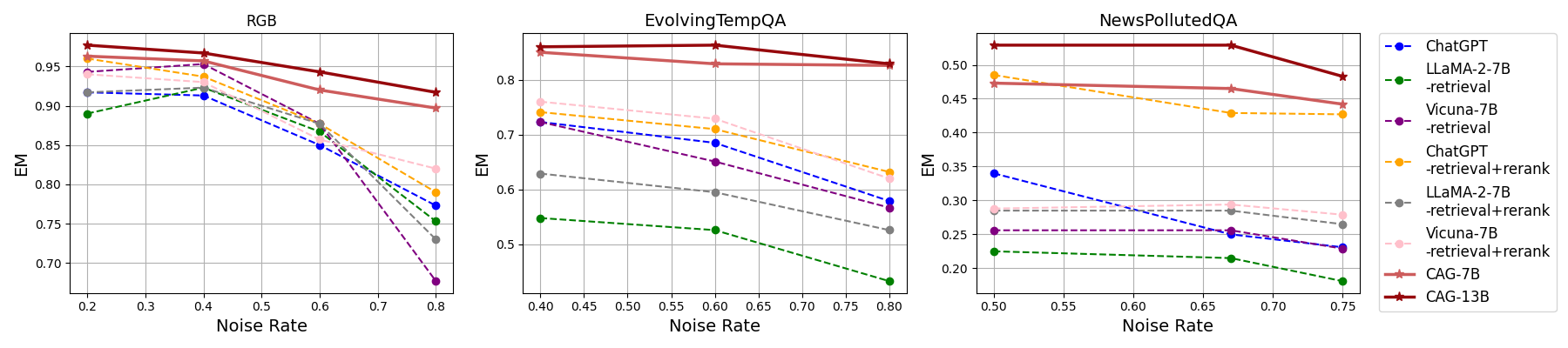}
    \caption{
    The performance of LLMs under varying noise ratios, which denote the proportions of retrieved noise documents.
    As the noise ratio increases, the performance of other methods markedly declines; in contrast, our model maintains stable performance in high noise ratio, attributed to its enhanced ability to prioritize accurate information.}
    \label{fig:trend}
\end{figure*}

\paragraph{Experimental settings}
We use LLaMA-2-7B, 13B and Mistral-7B as our base models. We train the LLaMA-2 model and the Mistral 7B model for 3 epochs with a learning rate of 1e-5 on A100-80G GPUs.To provide relevance scores, we use SPLADE \citep{Formal2021SPLADEVS} as our retriever. 
For all language models, we include 3-shot QA examples within the prompt. We utilize Exact Match (EM) \citep{stelmakh-etal-2022-asqa} as the primary evaluation metric for all datasets. It is calculated by checking whether the short answers provided are exact substrings of the generation.
We set the temperature to 0.01 during the inference.
Additional experimental settings and the prompts used for evaluation and are provided in the Appendix \ref{eval_prompt}. These models can generate answers with reasoning processes based on the given prompt. Moreover, Appendix \ref{appendix:cot} shows the results of the experiments using CoT prompt consistent with that used to generate training data.

\subsection{Overall Results}

The main results of the three scenarios are shown in the Table \ref{tab:results}, we can clearly see that our model efficiently utilizes credibility to provide more accurate and credible responses. In the following, we analyze the experimental results in detail:

\paragraph{1) Previous approaches based on RAG severely suffer from the flawed information introduced during retrieval.}
In scenarios including open-domain QA, time-sensitive QA, and misinformation pollutedQA, existing LLMs, including ChatGPT and LLaMA-2-70B, face challenges due to interference from flawed information. In the retrieval-based open-domain QA, the average EM score for ChatGPT is only 41.5\%, while 44.1\% for LLaMA-2-70B.
All models exhibit low performance on the Musique, NewsPollutedQA, which are characterized by high ratios of flawed information.
Reranking with external relevance scores can assist the model to a certain extent, as the model is sensitive to the order of documents \citep{xie_adaptive_2023}.

\paragraph{2) CAG significantly improves performance by discerning between documents and guiding the model to prioritize those with high credibility.} 
Our models significantly surpass all baseline models across the 7 datasets under 3 scenarios, including ChatGPT and LLaMA-2-70B enhanced with retrieval and reranking. For instance, on the 2WikiMHQA dataset, our CAG-7B improves 26.6\% of EM score over the LLaMA-2-7B model and 28.2\% of EM score over the Vicuna-7B model under retrieval-based.


\paragraph{3) Our approach generalizes to scenarios previously unseen which require credibility and demonstrates compatibility with diverse base models.}
The models, developed through training on LLaMA 7B, 13B, and Mistral 7B with CAG, not only exhibit improved reliability in its outputs but also excel in new, challenging situations, including time-sensitive QA and misinformation polluted QA. This performance, achieved within an open-domain QA framework lacking temporal or source integration, effectively manages diverse flawed information and affirms the universality of CAG.
\subsection{Analysis Study}
In the following, we will present analysis against the robustness and limitation of current CAG model. Due to the space limit, experimental results on the effect of credibility annotation accuracy are shown in the Appendix \ref{annotation_acc}.
\subsubsection{Noise Robustness Analysis}


Previous research has demonstrated that an increase in the proportion of noise within the context significantly degrades model performance \citep{xie_adaptive_2023,chen_benchmarking_2023}.
To assess the robustness of diverse methods against flawed information, we vary the ratio of noisy documents across three distinct datasets: RGB, EvolvingTempQA and NewsPollutedQA, and observe the consistency in performance changes across different models.

We present the results in Figure \ref{fig:trend} and can see that:
\textbf{Credibility-aware Generation makes the model robust to flawed information, which enhances its ability to discern and prioritize accurate information.}
As the proportions of noise in the context increases, most of the models exhibit performance degradation aligning with the observations made by \citet{chen_benchmarking_2023}. 
However, our models show greater robustness compared to others, notably the improved performance of CAG-13B on EvolvingTempQA. The results of the noise robustness analysis for all LLMs are shown in the Appendix \ref{appendix:noise}.

\subsubsection{Analysis of Discarding Low Credibility Documents}
\begin{figure}[t]
\setlength{\belowcaptionskip}{-0.5cm}
    \centering
\includegraphics[scale=0.3]{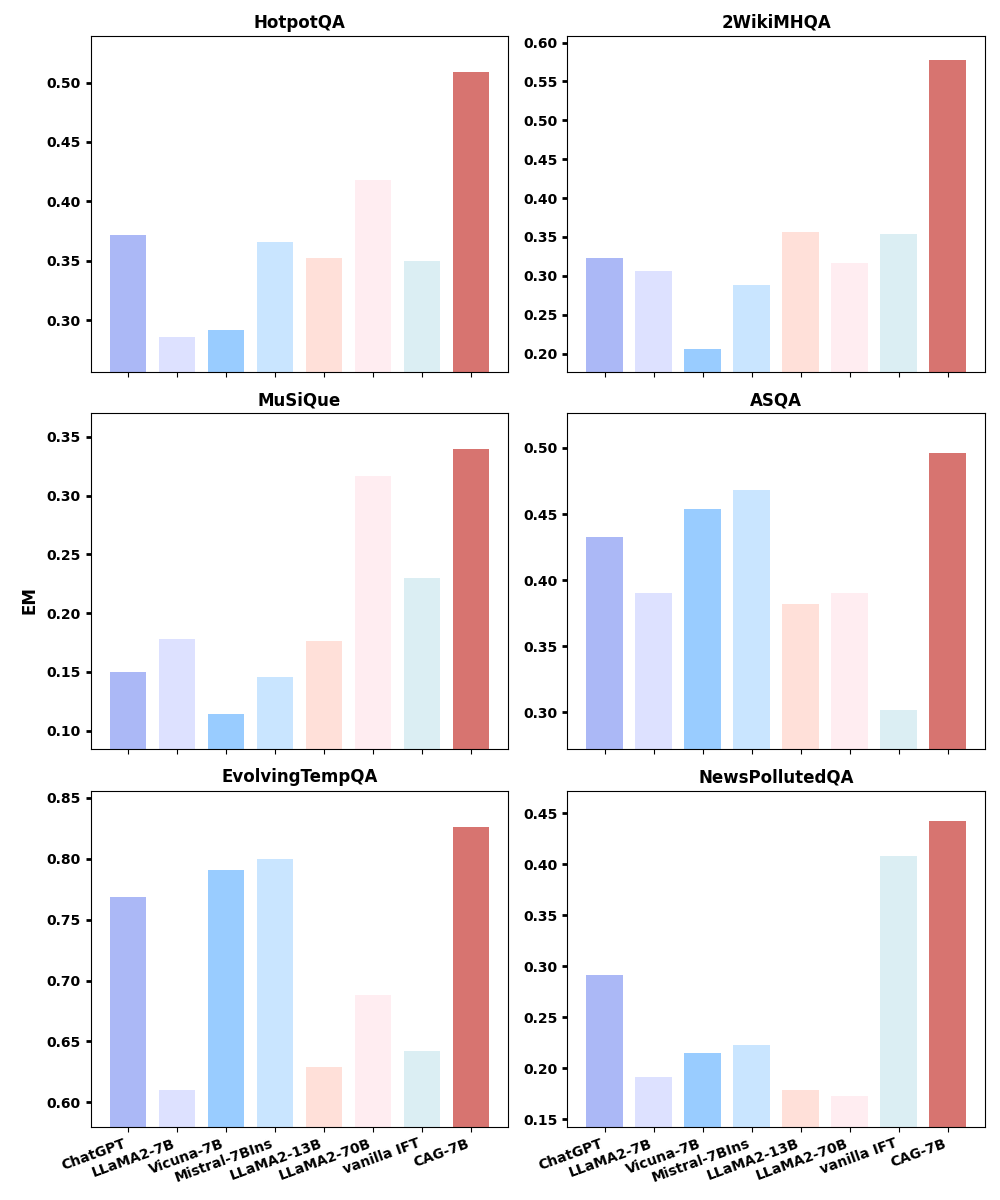}
    \caption{The comparison of performance of LLMs under discarding low credibility document setting and CAG-7B across six datasets.}
   \label{fig:discard}
\end{figure}
Upon assigning credibility to the documents in context, an alternative intuitive strategy is to simply discard low credibility documents. However, considering that credibility assessments are not precise, this strategy may inadvertently filter out helpful information, thereby impairing the accuracy of the model's responses.
To demonstrate this, we compare the performance of LLMs in this setting with that of CAG-7B in our CAGB.
The results are shown in Figure \ref{fig:discard}, and full results are in Appendix \ref{app:discard_full}. We can clearly see that:
by preserving more document information and differentiating them based on explicit credibility in the prompt, our CAG framework mitigates the risk of losing valuable information. As a result, the accuracy and comprehensiveness of the responses are improved.

\section{Related Work}
\label{related}

Retrieval-Augmented Generation~\citep{lewis2020retrieval} integrates a retriever with a generator to improve text generation quality by utilizing external knowledge \citep{izacard-grave-2021-leveraging, borgeaud2022improving, shi2023replug}. 
However, the accuracy of RAG is compromised by flawed information, as the inclusion of noisy \citep{chen_benchmarking_2023}, outdated \citep{kasai2022realtime,wang2023survey}, or false information 
 \citep{chen2023llmgenerated, pan2023risk} during the retrieval negatively impacts the generator's outputs.

Previous studies have primarily focused on filtering, ranking, or manually evaluating retrieved documents to mitigate the impact of flawed information.
For instance, \citet{peng_check_2023, wang_learning_2023} deploy various filtering algorithms to remove irrelevant text.
\citet{zhang2023mitigating} utilizes timestamps to identify and discard outdated information. 
However, these approaches are limited by the accuracy of filtering algorithms, thereby discarding helpful information and impairing the effectiveness of RAG.
Meanwhile, misinformation is primarily addressed by identifying falsehoods through fact-checking \citep{vijjali2020stage}. However, this approach necessitates either human verification or further training of the discriminator \cite{baek2023knowledgeaugmented}, both of which can be resource-intensive and introduce bias \citep{ Draws2022TheEO,su2023fake}.
In comparison, our work mitigates the impact of flawed information without discarding documents by introducing multi-feature dimensions of external information to assess the credibility level of each document.

Researchers fine-tune language models to better leverage the context provided in the input. For instance, \citet{li2023large} train the model using counterfactuals and irrelevant context to prioritize context. \citet{yoran2023making} include irrelevant context in the training samples, making the model robust to irrelevant documents. \citet{asai_self-rag_2023} train the model on contexts with reflective tokens, enabling it to evaluate the relevance of passages during generation.
However, these approaches focus mainly on irrelevant documents. Meanwhile, the model predominantly learns implicit rules, resulting in opaqueness of the generation.

\section{Conclusions}\label{conclusions}


This paper proposes Credibility-aware Generation to address the challenge of flawed information. To equip the model with CAG capabilities, we introduce a data transformation framework aimed at generating credibility-based dataset, upon which we fine-tune the model.
To effectively verify the ability of model Credibility-aware Generation to handle flawed information, we construct a benchmark from different real-world scenarios. Experimental results show that our model can effectively utilize credibility, exhibiting robustness in the face of flawed information and significantly outperforming other models with retrieval augmentation.

Moreover, through customizing the credibility, our approach can be applied to the real-world scenario including personalized response generation, for which we provide a detailed case study in the Appendix \ref{case}.
\section*{Limitations}
There are several limitations of our current CAG framework, which we plan to address in the future. 
Firstly, we have established a flexible credibility assessment mechanism, focusing more on endowing the model with the ability to generate based on credibility. However, credibility assessment is also a crucial part, and the current performance gap exists due to the retrieval strategy and influencing factors. In future research, we will delve further into credibility assessment to enhance the performance of our model.
Secondly, despite our method demonstrating strong generalization capabilities, it still relies on additional training data annotation and training. In the future, we will explore how to enable existing models to perform confidence-aware generation without the need for further training.
Thirdly, our methodology, effectively applied to RAG, acknowledges the broader research domain encompassing external resources like knowledge graphs and tool usage. We aim to expand our work to domains requiring diverse external information integration, including retrieved data, knowledge graph data, and tool output.


\section*{Acknowledgements}
We sincerely thank the reviewers for their insightful comments and valuable suggestions. This work was supported by Beijing Natural Science Foundation (L243006), the Natural Science Foundation of China (No. 62122077 and 62106251), and Beijing Municipal Science and Technology Project (Nos. Z231100010323002).
\bibliography{anthology_00,custom,anthology_01}
\bibliographystyle{acl_natbib}
\clearpage
\appendix

\section{Appendix}
\label{sec:appendix}
\subsection{Overview of Training Data Statistics \label{train_data}}
The composition and statistics of the training data are as follows:
\begin{table}[h]\small
\centering
\resizebox{0.48\textwidth}{!}{\begin{tabular}{lll}
\toprule
\textbf{Task} & \textbf{Dataset}  & \textbf{Train (\#)} \\ \midrule
Dialogue & ShareGPT  \citep{vicuna2023} & 3426 \\
\multirow{4}{*}{ODQA} & HotpotQA\citep{yang-etal-2018-hotpotqa} &  5287      \\
& ELI5 \citep{fan-etal-2019-eli5} &  2000    \\
& QAMPARI\citep{amouyal2023qampari} &   1000   \\
& WikiQA\citep{yang2015wikiqa} &   1040    \\
\multirow{1}{*}{MRC} & NewsQA\citep{trischler-etal-2017-newsqa} & 2135 \\
 & PubmedQA\citep{jin-etal-2019-pubmedqa} &   12552    \\
\bottomrule
\end{tabular}
}
\caption{\label{tab:train_data} Statistics of our training data with multiple-granularity credibility annotation and credibility-guided explanation.}
\end{table}
\subsection{Effect of Credibility Annotation Accuracy \label{annotation_acc}}
To investigate the impact of credibility annotation accuracy on the performance of CAG and to identify the upper limit of their potential, 
We conduct a comparison between the use of golden credibility annotations and retriever-based credibility annotations within open-domain QA using both the CAG-7B and CAG-13B models. Golden credibility annotations refer to labeling golden support evidence as high credibility and other text as low credibility.

The results of our experiments are presented in Table \ref{tab:golden}.
We can find that: The precision of retrieval model annotation credibility is a primary factor limiting the current performance of CAG.
The results, as presented, clearly demonstrate that reliable credibility annotations are instrumental in unlocking the model's potential. Compared with the use of SPLADE to label credibility, the use of golden credibility labels on the CAG-7B has resulted in an average improvement of 14.4\% of EM across three datasets.

\begin{table}[h]
\setlength{\belowcaptionskip}{-0.5cm}
\centering
\resizebox{0.4\textwidth}{!}{
\begin{tabular}{cccc}
\toprule
\textbf{Dataset}& \textbf{Annotation} & \textbf{CAG-7B} & \textbf{CAG-13B} \\
\midrule
\multirow{2}{*}{2WikiMHQA} & SPLADE  & 0.562 & 0.604 \\
 & Golden & \textbf{0.698} & \textbf{0.650}\\ 
\midrule
\multirow{2}{*}{Musique} & SPLADE & 0.340 & 0.408 \\
 & Golden & \textbf{0.626} & \textbf{0.656}\\
\midrule
\multirow{2}{*}{ASQA} & SPLADE & 0.496 & 0.510\\
 & Golden & \textbf{0.505} & \textbf{0.525}\\
 \midrule
\multirow{2}{*}{Average} & SPLADE & 0.466 & 0.507\\
 & Golden & \textbf{0.610} & \textbf{0.610}\\
\bottomrule
\end{tabular}
}
\caption{\label{tab:golden}
The performance comparison of the CAG-7B and CAG-13B when using retrieved annotation credibility and golden credibility annotations.}
\end{table}
\newpage
\subsection{Fine-grained Credibility Analysis\label{fine-grain}}

To investigate the performance differences between fine-grained credibility and the three-level credibility method we currently use, we select several representative models and datasets. The fine-grained credibility employed ranges from a credibility score of 0 to 9, based on relevance. The experimental results are presented in Table \ref{tab:fine}.
We can see that the use of fine-grained credibility models may lead to a decrease in performance. Fine-grained credibility demands higher accuracy in credibility classification and greater capability from the model to understand and differentiate credibility levels.
\begin{table}[h]
    \centering
    \resizebox{0.45\textwidth}{!}{\begin{tabular}{lllll}
        \toprule
        \textbf{Model}  & \textbf{HotpotQA} & \textbf{2WikiMHQA} & \textbf{MuSiQue} &
        \textbf{EvolvingTempQA}\\
        \midrule
        ChatGPT & 39.1 {\color{teal}$\downarrow$}3.1 & 36.0 {\color{teal}$\downarrow$} 4.2& 23.6 {\color{red}$\uparrow$}5.4 & 66.0 {\color{teal}$\downarrow$}1.3\\
        Vicuna-7B  & 27.9 {\color{teal}$\downarrow$}7 & 28.4 {\color{red}$\uparrow$} 1.8& 11.4 {\color{red}$\uparrow$}2.3 & 62.4 {\color{teal}$\downarrow$}1.8
        \\
        LLaMA-2-7B & 28.5 {\color{teal}$\downarrow$} 9.1& 26.4{\color{red}$\uparrow$}8.8  & 13.4 {\color{teal}$\downarrow$} 0.6 & 45.3{\color{teal}$\downarrow$} 3.3\\
        LLaMA-2-13B  & 34.1 {\color{teal}$\downarrow$}1.9 & 33 {\color{teal}$\downarrow$} 5.4& 15{\color{teal}$\downarrow$}1.4& 49.9{\color{teal}$\downarrow$}2.1\\
        \bottomrule
    \end{tabular}}
    \caption{Performance of models using fine-grained credibility. The number following the downward arrow indicates the performance degradation compared to the currently used credibility granularity.}
    \label{tab:fine}
\end{table}
\subsection{Retain the documents with the highest similarity}
We conduct experiments on multi-hop QA datasets under the setting that only the most similar documents are retained.
 Based on the number of documents required to answer questions in each dataset, we retain the top 2 documents for HotpotQA and 2WikiMHQA, and the top 5 documents for the MuSiQue dataset. 
 We compare the performance of the model under this strategy with our CAG-7B model, as shown in Table \ref{tab:sim}. The experimental results indicate that discarding the majority of low-similarity texts may enhance model performance. However, it still does not surpass our model, which retains as much information as possible while minimizing interference from irrelevant information in the presence of high credibility documents. Additionally, relying solely on low- similarity filtering is inadequate for removing outdated and false information.
\begin{table}[H]
    \centering
    \resizebox{0.45\textwidth}{!}{\begin{tabular}{llll}
        \toprule
        \textbf{Model}  & \textbf{HotpotQA} & \textbf{2WikiMHQA} & \textbf{MuSiQue}\\
        \midrule
        ChatGPT & 0.398  & 0.318 & 0.150\\
        Vicuna-7B  & 0.353 & 0.284 & 0.174\\
        LLaMA-2-13B  & 0.375 & 0.408 & 0.228\\
        \midrule
        CAG-7B & \textbf{0.509} & \textbf{0.578} & \textbf{0.340} \\
        \bottomrule
    \end{tabular}}
    \caption{Performance of models under discarding most low-similarity documents.}
    \label{tab:sim}
\end{table}
\subsection{Customized Credibility Applications\label{case}} 
In demonstrating the capability of customized credibility in CAG, this paper presents 3 examples that highlight its diverse application scenarios, including personalized response generation and the resolution of knowledge conflicts.

\subsubsection{Personalized Response Generation \label{case1}}
\begin{figure}[H]
\setlength{\belowcaptionskip}{-0.05cm}
    \centering
    \subcaptionbox{Based on user search history, CAG generates personalized and targeted responses.\label{case1}}{
        \includegraphics[width = .48\textwidth]{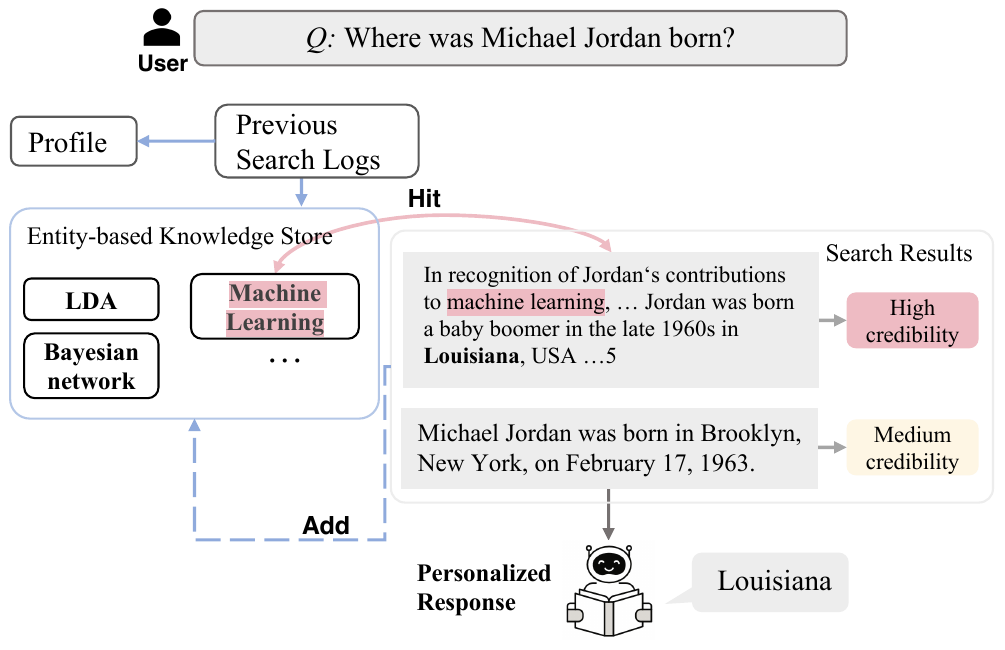}
    }
    \subcaptionbox{CAG provides personalized destination recommendations based on user profile.\label{case2}}{
        \includegraphics[width = .48\textwidth]{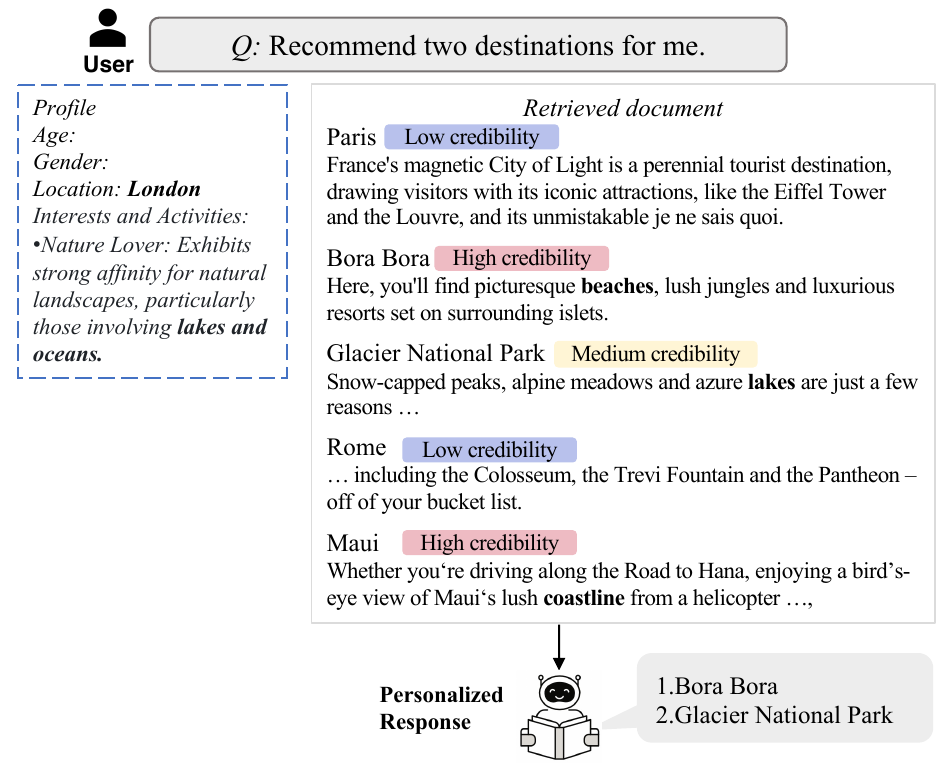}
    }
    \caption{CAG provides personalized responses. 
We can see that CAG combines with user preferences to utilize customized credibility, offering personalized responses.\label{fig:case1}}
\end{figure}

LLMs tailored to individuals consider individual preferences and requirements, thereby enhancing service precision and user satisfaction.

\citet{baek2024knowledgeaugmented} maintain an entity-centric knowledge base from the user's search history, enriching LLM to provide customized services. This knowledge base reflects users' current and potential interests. Upon receiving a novel query, the system initially retrieves relevant content. If the obtained entities correspond to those present in the user's knowledge base, the system deems this information relevant, attributing higher credibility to the associated documents. Consequently, the CAG module 
 can generate personalized responses based on documents with credibility annotations, as illustrated in Figure \ref{case1}. Moreover, by maintaining user profiles to record preference,in recommendation scenarios, the system retrieves numerous documents based on user input and assigns credibility to documents based on their alignment with the user's profile, achieving personalized and controllable recommendations, as show in Figure \ref{case2}.

\subsubsection{Knowledge Conflict Resolution}

\begin{figure}[H]
    \centering
    \includegraphics[scale=0.43]{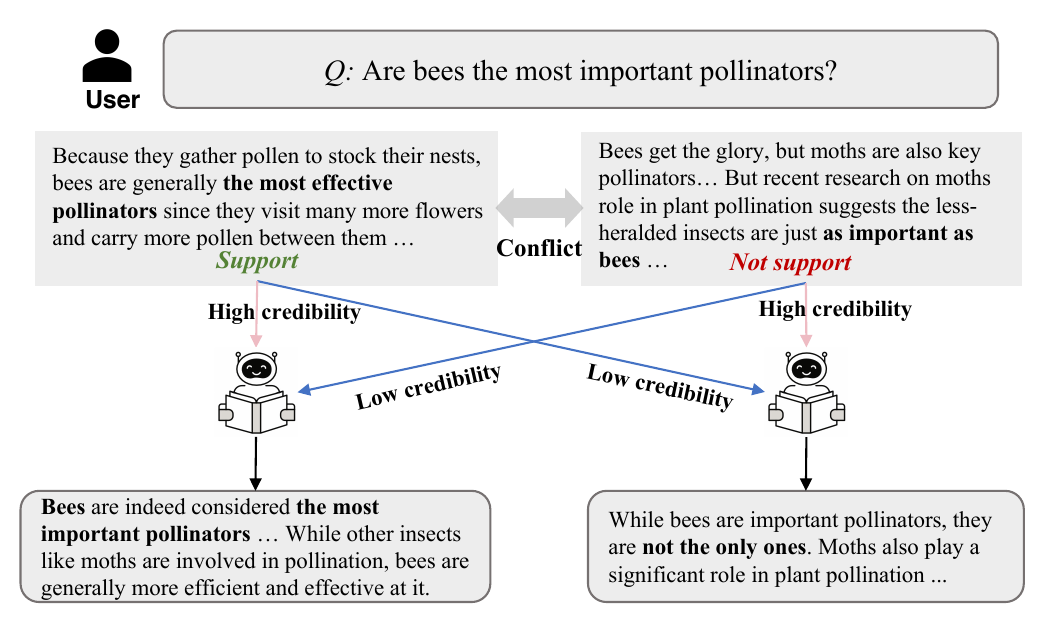}
    \caption{By assigning documents different credibility degrees, CAG resolves knowledge conflicts.}
    \label{fig:case3}
\end{figure}
In real-world scenarios, controversial questions are often encountered, and the retrieved documents tend to contain contradictory evidence. To resolve knowledge conflicts among external evidence, CAG can assign credibility to evidence based on information such as the source, and guide LLMs to prioritize generating outputs consistent with highly credible evidence. Figure \ref{fig:case3} illustrates a simple example, where the sample question comes from a dataset specifically focused on controversial issues in real-world scenarios \cite{wan2024evidence}. Therefore, CAG can be utilized to resolve conflicts between public databases and private data, as well as between general knowledge bases and proprietary knowledge bases, by assigning high credibility to private data and proprietary knowledge bases.

\newpage
\subsection{ASQA Full Results \label{appendix:asqa}}
Table \ref{tab:asqa} shows all results of LLMs on ASQA.
\begin{table}[H]
\centering
\resizebox{0.4\textwidth}{!}{
\begin{tabular}{lccc}
\toprule
\textbf{Model} & \textbf{Length} & \textbf{EM} & \textbf{Rouge-L} \\
\midrule
\rowcolor{graybg} \multicolumn{4}{c}{\textit{retrieval-based}}\\ \midrule
ChatGPT & 0.400$^*$ & 0.404$^*$ & 0.370$^*$ \\
LLaMA-2-7B & 41.6 & 26.8 & 31.0 \\
Vicuna-7B-v1.5 & 65.4 & 35.8 & 36.6 \\
Mistral-7B-Instruct & 25.7 & 30.0 & 34.0 \\
LLaMA-2-13B & 30.7 & 32.1 & 33.6 \\
LLaMA-2-70B & 16.1 & 31.6 & 31.6 \\
vanilla IFT & 23.7 & 15.7 & 23.1 \\
\midrule
\rowcolor{graybg} \multicolumn{4}{c}{\textit{retrieval and reranking}}\\ \midrule
ChatGPT & 40.8$^*$ & 40.2$^*$ & 36.9$^*$ \\
LLaMA-2-7B & 38.1 & 37.5 & 32.5 \\
Vicuna-7B-v1.5 & 66.1 & 49.4 & 38.5 \\
Mistral-7B-Instruct & 24.5 & 41.4 & 35.7 \\
LLaMA-2-13B & 30.0 & 39.0 & 34.9 \\
LLaMA-2-70B & 16.3 & 38.8 & 33.0 \\
vanilla IFT & 23.8 & 17.6 & 23.0 \\
\midrule
\rowcolor{graybg} \multicolumn{4}{c}{\textit{retrieval and credibility}}\\ \midrule
ChatGPT & 30.4 & 44.0 & 38.5 \\
LLaMA-2-7B & 54.2 & 39.4 & 34.2 \\
Vicuna-7B & 64.9 & 49.0 & 38.5 \\
Mistral-7B-Instruct & 52.3 & 46.3 & 39.2 \\
LLaMA-2-13B & 39.1 & 38.5 & 33.6 \\
LLaMA-2-70B & 49.6 & 49.2 & 39.7 \\
vanilla IFT & 3.4 & 30.5 & 9.2 \\ \hline
CAG-7B & 94.0 & 50.3 & 39.3 \\
CAG-13B & 80.4 & 52.5 & 40.3 \\
CAG-mistral-7B & 69.7 & 50.5 & 40.3 \\
\bottomrule
\end{tabular}
}
\caption{All results of LLMs on ASQA. The results of EM and Rouge-L are displayed multiplied by 100. $^*$ indicates result reported from \citet{gao2023enabling}.}
\label{tab:asqa}
\end{table}
\subsection{Experimental Results Using CoT prompt}
\label{appendix:cot}
\begin{figure}[H]
    \centering
    \includegraphics[scale=0.38]{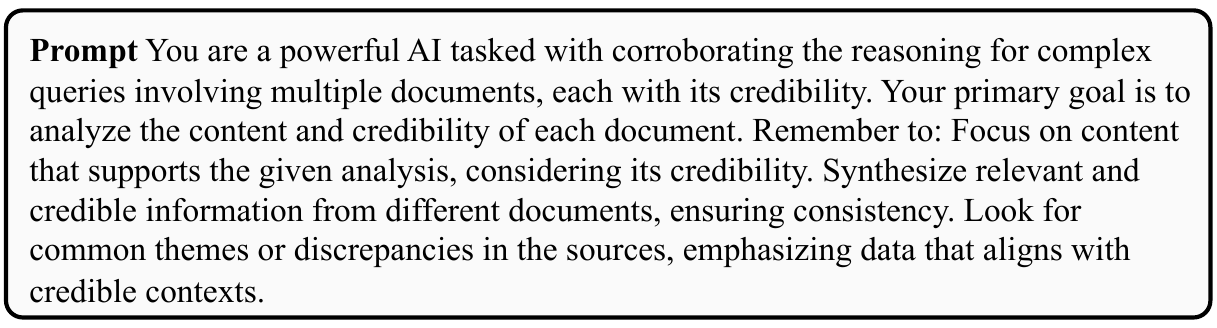}
    \caption{The CoT prompt used for evaluation.}
    \label{fig:cot}
\end{figure}

Figure \ref{fig:cot} shows the CoT prompt which is consistent with the generation process of training data.
Table \ref{tab:cot} demonstrates the experimental results of the models using CoT prompt on CAGB. 
\begin{table*}[h]
\centering
\resizebox{0.8 \textwidth}{!}{\begin{tabular}{lccccccc}
\toprule
   \textbf{Model} &\textbf{HotpotQA}& \textbf{2WikiMHQA} & \textbf{MuSiQue} & \textbf{ASQA}    & \textbf{RGB}  & \textbf{EvolvingTempQA}    &  \textbf{NewsPollutedQA}  \\
\midrule
ChatGPT+CoT & 0.46 & 0.398 & 0.262 & 0.448 & 0.883 & 0.806 & 0.450 \\
Vicuna-7B+CoT & 0.316 & 0.506 & 0.202 & 0.479 & 0.823 & 0.766 & 0.392 \\
Mistral-7B-Instruct+CoT & 0.392 & 0.314 & 0.102 & 0.475 & 0.74 & 0.752 & 0.394 \\
LLaMA-2-13B+CoT & 0.336 & 0.33 & 0.158 & 0.389 & 0.78 & 0.478 & 0.273 \\
LLaMA-2-70B+CoT & 0.382 & 0.432 & 0.254 & 0.480 & 0.837 & 0.58 & 0.298 \\
CAG-7B (ours) & 0.509 & 0.578 & 0.340 & 0.496 & 0.897 & 0.826 & 0.442 \\
CAG-13B (ours) & 0.514 & 0.604 & 0.408 & 0.525 & 0.917 & 0.829 & 0.483 \\
\bottomrule
\end{tabular}}
\caption{Experimental results of models using CoT prompt on the CAGB.}
\label{tab:cot}
\end{table*}
\subsection{Full Results Under the Discarding Low Credibility Documents \label{app:discard_full}}
Table \ref{tab:discard_full} shows the full results under the \textit{discarding low credibility documents} setting.
\begin{table*}[h]
    \centering
    \resizebox{0.8\textwidth}{!}{
    \begin{tabular}{lccccccc}
    \toprule
   \textbf{Model} &\textbf{HotpotQA}& \textbf{2WikiMHQA} & \textbf{MuSiQue} & \textbf{ASQA}    & \textbf{RGB}  & \textbf{EvolvingTempQA}    &  \textbf{NewsPollutedQA}  \\ \midrule 
ChatGPT          & 0.372   & 0.323   & 0.150   & 0.433   & 0.760     & 0.769  & 0.291      \\
LLaMA-2-7B       & 0.286   & 0.306   & 0.178   & 0.390   & 0.710     & 0.610  & 0.192     \\
Vicuna-7B        & 0.292   & 0.206   & 0.114   & 0.454   & 0.737     & 0.791  & 0.215    \\
Mistral-7B-Instruct & 0.366 & 0.288  & 0.146   & 0.468 & 0.757 & 0.800 & 0.223 \\
LLaMA-2-13B      & 0.352    & 0.356   & 0.176   & 0.382   & 0.783     & 0.629  & 0.179    \\
LLaMA-2-70B      & 0.418   & 0.316   & 0.317   & 0.477   & 0.840 & 0.688  & 0.173    \\
vanilla IFT      & 0.350   & 0.354   & 0.230   & 0.302   & 0.723& 0.642& 0.408    \\
CAG-7B (ours) & 0.509 & 0.578 & 0.340 & 0.496 & 0.897 & 0.826 & 0.442 \\
CAG-13B (ours) & 0.514 & 0.604 & 0.408 & 0.525 & 0.917 & 0.829 & 0.483 \\
\bottomrule
    \end{tabular}
    }
    \caption{Full experimental results under the \textit{discarding low credibility documents} setting.}
    \label{tab:discard_full}
\end{table*}

\subsection{Details of Credibility Assessment \label{cred_asses}}
\begin{figure}[h]
    \centering
    \includegraphics[scale=0.6]{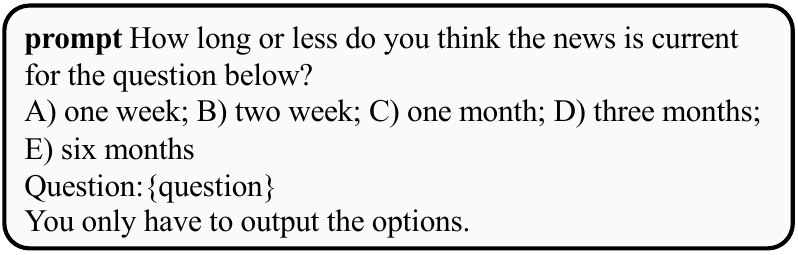}
    \caption{The prompt used to evaluate the validity period.}
    \label{fig:time}
\end{figure}

The process of credibility assessment also encompasses the determination of a temporal threshold.
The method we employ is designing prompts that allow the LLM to assess the timeliness of news articles regarding the question within varying temporal scopes. This approach takes into account the inherent validity period of the events within the question. In order to ensure the stability of the validity period evaluation, we conduct three trials, voting to select the validity period within each question. The prompt that we design can be found in Figure \ref{fig:time}.
\subsection{A Comparison of CAGB with Other Similar Benchmarks}
\begin{table}[H]\small
\resizebox{0.45\textwidth}{!}{\begin{tabular}{lllll}
\toprule
& Noise Info & Outdated Info & Misinfo & Golden Annotation \\
\midrule
KILT & \textcolor{olivegreen}{\Checkmark} & \XSolidBrush & \XSolidBrush & \textcolor{olivegreen}{\Checkmark} \\
RealTime QA & \textcolor{olivegreen}{\Checkmark} & \textcolor{olivegreen}{\Checkmark} & \XSolidBrush & \XSolidBrush \\
Streaming QA & \textcolor{olivegreen}{\Checkmark} & \textcolor{olivegreen}{\Checkmark} & \XSolidBrush & \XSolidBrush \\
Misinfo QA & \XSolidBrush & \XSolidBrush & \textcolor{olivegreen}{\Checkmark} & \textcolor{olivegreen}{\Checkmark} \\
CAGB (\textit{ours}) & \textcolor{olivegreen}{\Checkmark} & \textcolor{olivegreen}{\Checkmark} & \textcolor{olivegreen}{\Checkmark} & \textcolor{olivegreen}{\Checkmark} \\
\bottomrule
\end{tabular}}
\caption{Comparison with existing benchmarks. \label{tab:benchmark}}
\end{table}

\subsection{Prompts Used on the CAGB \label{eval_prompt}}
\begin{figure}[H]
    \centering
    \includegraphics[scale=0.6]{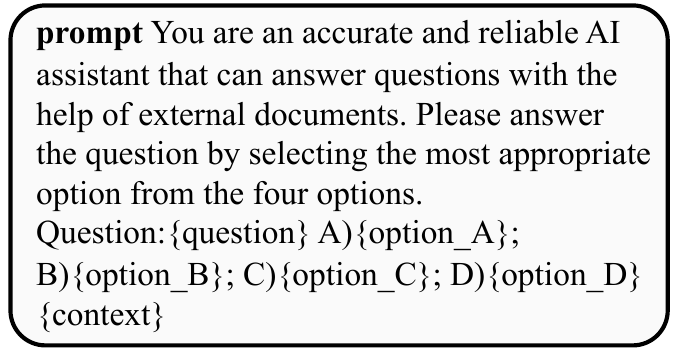}
    \caption{The prompt used in the retrieval-based and retrieval and reranking settings on the NewsPollutedQA dataset.}
    \label{fig:mis1}
\end{figure}

\begin{figure}[h]
    \centering
    \includegraphics[scale=0.48]{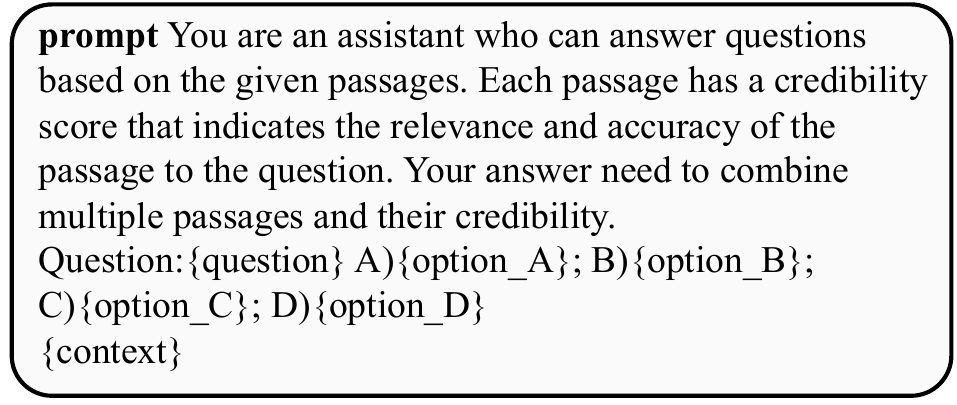}
    \caption{The prompt used in the retrieval-based and retrieval and reranking settings on the NewsPollutedQA dataset.}
    \label{fig:mis2}
\end{figure}

\begin{figure}[H]
    \centering
    \includegraphics[scale=0.55]{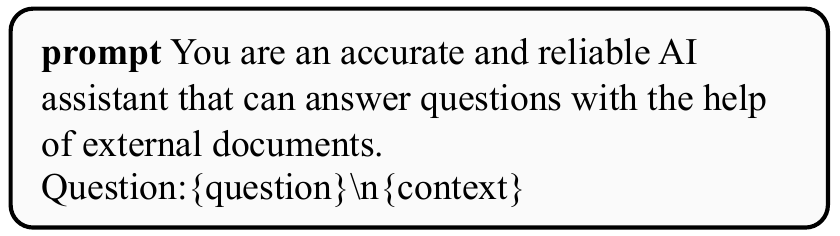}
    \caption{The prompt used in the retrieval-based and retrieval and reranking settings.}
    \label{fig:retrieve}
\end{figure}

\begin{figure}[t]
    \centering
    \includegraphics[scale=0.55]{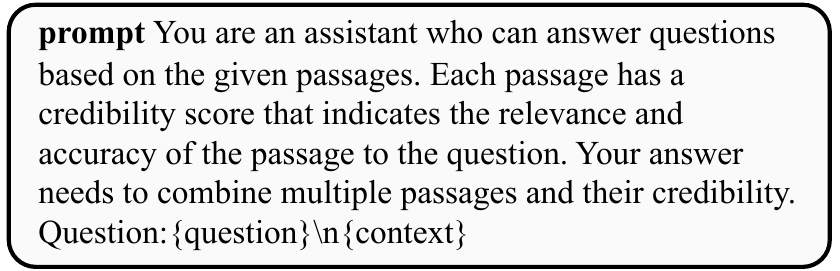}
    \caption{The prompt used in the retrieval and credibility settings.}
    \label{fig:cred}
\end{figure}

We conduct an evaluation of ASQA utilizing the prompts provided in \citet{gao2023enabling}. The prompts utilized for the evaluation of the NewsPollutedQA dataset, under the settings of retrieval-based, retrieval and reranking, and retrieval and credibility in the zero-shot scenario, are displayed in Figures \ref{fig:mis1} and \ref{fig:mis2}. The prompts used for assessing other datasets, under the settings of retrieval-based, retrieval and reranking, and retrieval and credibility in the zero-shot scenario, can be found in Figure \ref{fig:retrieve} and Figure \ref{fig:cred}.

\subsection{Prompt Used to Generate Credibility-guided Explanation \label{gen_prompt}}
\begin{figure}[]
    \centering
\includegraphics[scale=0.5]{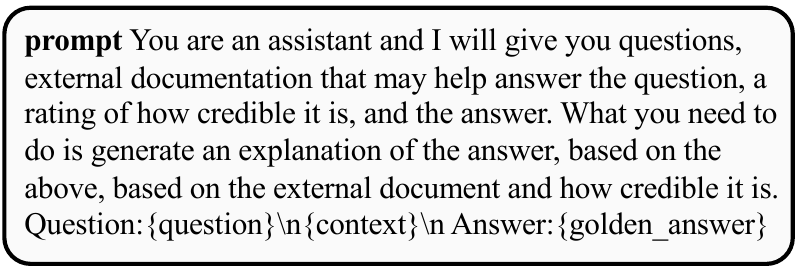}
    \caption{Prompt used to generate credibility-guided explanation.}
    \label{fig:gen}
\end{figure}

To guide the LM to credibility-guided explanation, we design the following prompt, as shown in Figure \ref{fig:gen}.
\newpage
\subsection{Results of the Noise Robustness Analysis \label{appendix:noise}}

Table \ref{tab:noise} presents the experimental results of the LLMs in noise ratio analysis on the RGB.

\begin{table}[H]
\resizebox{0.46\textwidth}{!}{
\begin{tabular}{lcccc}
\toprule
\multirow{2}{*}{\textbf{Model}}& \multicolumn{4}{c}{\textbf{Noise Ratio}} \\
 & 0.2 & 0.4 & 0.6 & 0.8 \\
\midrule
\rowcolor{graybg} \multicolumn{5}{c}{\textit{retrieval-based}}\\ \midrule
ChatGPT & 0.917 & 0.913 & 0.850 & 0.773 \\
LLaMA-2-7B & 0.890 & 0.890 & 0.877 & 0.753 \\
Vicuna-7B-v1.5 & 0.943 & 0.953 & 0.877 & 0.677 \\
LLaMA-2-13B & 0.903 & 0.907 & 0.870 & 0.820 \\
LLaMA-2-70B & 0.960 & 0.937 & 0.910 & 0.823 \\
vanilla IFT & 0.793 & 0.793 & 0.767 & 0.650 \\
Mistral-7B-Instruct & 0.900 & 0.903 & 0.880 & 0.713 \\
\midrule
\rowcolor{graybg} \multicolumn{5}{c}{\textit{retrieval and reranking}}\\ \midrule
ChatGPT & 0.960 & 0.937 & 0.877 & 0.790 \\
LLaMA-2-7B & 0.917 & 0.923 & 0.877 & 0.730 \\
Vicuna-7B-v1.5 & 0.940 & 0.930 & 0.857 & 0.820 \\
LLaMA-2-13B & 0.933 & 0.933 & 0.897 & 0.823 \\
LLaMA-2-70B & 0.957 & 0.960 & 0.927 & 0.833 \\
vanilla IFT & 0.833 & 0.780 & 0.767 & 0.663 \\
Mistral-7B-Instruct & 0.913 & 0.907 & 0.877 & 0.790 \\
\midrule
\rowcolor{graybg} \multicolumn{5}{c}{\textit{retrieval and credibility}}\\ \midrule
ChatGPT & 0.973 & 0.943 & 0.893 & 0.807 \\
LLaMA-2-7B & 0.903 & 0.917 & 0.877 & 0.713 \\
Vicuna-7B-v1.5 & 0.950 & 0.947 & 0.870 & 0.740 \\
LLaMA-2-13B & 0.920 & 0.910 & 0.897 & 0.803 \\
LLaMA-2-70B & 0.953 & 0.950 & 0.900 & 0.817 \\
vanilla IFT & 0.827 & 0.773 & 0.710 & 0.643 \\
Mistral-7B-Instruct & 0.940 & 0.910 & 0.867 & 0.797 \\ \hline
CAG-7B & 0.963 & 0.957 & 0.920 & 0.897 \\
CAG-13B & 0.977 & 0.967 & 0.943 & 0.917 \\
CAG-mistral-7B & 0.980 & 0.963 & 0.937 & 0.900 \\
\bottomrule
\end{tabular}
}
\caption{The performance of the LLMs under varying noise ratio on the RGB.}
\label{tab:noise}
\end{table}

Table \ref{tab:noise1} presents the experimental results of the LLMs in noise ratio analysis on the EvolvingTempQA, and NewsPollutedQA.
\begin{table}[t]
\centering
\resizebox{0.43 \textwidth}{!}{
\begin{tabular}{lcccccc}
\toprule
\multirow{3}{*}{\textbf{Model}}& \multicolumn{3}{c}{\textbf{EvolvingTempQA}} & \multicolumn{3}{c}{\textbf{NewsPollutedQA}}\\
& \multicolumn{3}{c}{Noise Ratio} & \multicolumn{3}{c}{Noise Ratio} \\
& 0.4 & 0.6 & 0.8 & 0.5 & 0.67 & 0.75 \\
\midrule
\rowcolor{graybg} \multicolumn{7}{c}{\textit{retrieval-based}}\\ \midrule
ChatGPT& 0.723 & 0.685 & 0.579 & 0.340         & 0.250         & 0.231         \\
LLaMA-2-7B      & 0.548 & 0.526 & 0.433 & 0.225         & 0.215         & 0.181         \\ 
Vicuna-7B-v1.5  & 0.723 & 0.651 & 0.567 &  0.256         & 0.256         & 0.229 \\
LLaMA-2-13B     & 0.645 & 0.579 & 0.495 & 0.263         & 0.267         & 0.204 \\
LLaMA-2-70B     &  0.651 & 0.586 & 0.526 & 0.277         & 0.254         & 0.192 \\
vanilla IFT & 0.667 & 0.651 & 0.592 &  0.463         & 0.452         & 0.369 \\
Mistral-7B-Instruct& 0.769 & 0.701 & 0.598 &  0.392         & 0.283         & 0.204\\
\midrule
\rowcolor{graybg} \multicolumn{7}{c}{\textit{retrieval and reranking}}\\ \midrule
ChatGPT& 0.741 & 0.710 & 0.632 & 0.485         & 0.429         & 0.427 \\ 
LLaMA-2-7B      & 0.629 & 0.595 & 0.526 & 0.285         & 0.285         & 0.265\\ 
Vicuna-7B-v1.5  & 0.760 & 0.729 & 0.620 & 0.283         & 0.296         & 0.275\\ 
LLaMA-2-13B     &  0.654 & 0.636 & 0.561 & 0.335         & 0.335         & 0.308\\ 
LLaMA-2-70B     & 0.664 & 0.620 & 0.570 & 0.423         & 0.396         & 0.306\\ 
vanilla IFT & 0.779 & 0.773 & 0.720 & 0.488         & 0.463         & 0.356\\ 
Mistral-7B-Instruct& 0.826 & 0.801 & 0.741 & 0.513         & 0.454         & 0.373\\ 
\midrule
\rowcolor{graybg} \multicolumn{7}{c}{\textit{retrieval and credibility}}\\ \midrule
ChatGPT& 0.773 & 0.757 & 0.673 & 0.604         & 0.588         & 0.408\\ 
LLaMA-2-7B      & 0.570 & 0.545 & 0.486 & 0.254         & 0.254         & 0.213\\ 
Vicuna-7B-v1.5  & 0.782 & 0.791 & 0.642 & 0.288         & 0.294         & 0.279\\ 
LLaMA-2-13B     & 0.639 & 0.607 & 0.520 & 0.325         & 0.310         & 0.227\\ 
LLaMA-2-70B     & 0.673 & 0.645 & 0.611 & 0.471         & 0.400         & 0.279\\ 
vanilla IFT &  0.685 & 0.657 & 0.589&0.481 & 0.477& 0.427 \\ 
Mistral-7B-Instruct& 0.804 & 0.773 & 0.679& 0.515         & 0.402         & 0.315\\ \hline
CAG-7B & 0.850 & 0.829 & 0.826 & 0.473         & 0.465         & 0.442\\ 
CAG-13B& 0.860 & 0.863 & 0.829 & 0.529         & 0.529         & 0.483\\ 
CAG-mistral-7B     & 0.832 & 0.844 & 0.835 & 0.679         & 0.640         & 0.613\\ 
\bottomrule
\end{tabular}
}
\caption{The performance of the LLMs under varying noise ratio on the EvolvingTempQA and NewsPollutedQA.}
\label{tab:noise1}
\end{table}
\newpage
\subsection{Examples of CAGB}
Table \ref{app:evo} and \ref{app:news} present some examples of CAGB.
\begin{table*}[t]
    \centering
    \begin{tabular}{>{\raggedright\arraybackslash}p{11cm} >{\raggedright\arraybackslash}p{3cm}}
        \toprule
        \textbf{Input} & \textbf{Answer} \\
        \midrule
        \textcolor{olivegreen}{Question:}More than 30,000 pounds of which food product were recently recalled? date:2011/10/23 & Tyson Foods \\
        \textcolor{olivegreen}{Docs:}\textcolor{red}{High credibility of text:} Tyson Foods is voluntarily recalling almost 30,000 pounds of its dinosaur-shaped chicken nuggets due to possible contamination of foreign materials, specifically metal pieces, according to a press release issued by the U.S. Department of Agriculture's Food Safety and Inspection Service on Saturday. date:2023/11/06 & \\
        \textcolor{cyan}{Low credibility of text:} Washington Beef recalls 30,000 pounds of product:The FDA announced a large recall for Washington Beef products that could contain hard plastic or metal. date:2019/03/06 \\
        \textcolor{cyan}{Low credibility of text:} Perdue Foods recalls 30k pounds of chicken products: Perdue Foods, LLC. recalled more than 30,000 pounds of ready-to-eat chicken products after consumer complaints were received, according to the United States Department of Agriculture's Food Safety and Inspection Service. The products may contain, "extraneous materials, specifically pieces of bone," according to a release by the agency.The recall was classified as 'Class I,' meaning there is a, "reasonable probability that the use of the product will cause serious, adverse health consequences or death." However, there have been no confirmed reports of adverse reactions. date:2019/06/02 \\
        \textcolor{red}{High credibility of text:} Tyson Foods is recalling nearly 30,000 pounds of breaded chicken "Fun Nuggets" after consumers complained of finding metal pieces in the dinosaur-shaped patties. The nuggets,sold in 29-ounce bags,were produced on Sept. 5 by the Berryville,Arkansas,company. Tyson informed the U.S. Department of Agriculture's Food Safety and Inspection Service and said it recalled the nuggets voluntarily "out of an abundance of caution." USDA said that there had been only one report of a "minor oral injury associated with consumption of this product." date:2023/11/06 \\
        \textcolor{cyan}{Low credibility of text:} 30,000 Pounds of Beef Recalled Over Contamination Fears The Food Safety and Inspection Service issued a Class I recall for certain products from Washington Beef,indicating that ‘severe’ health consequences could occur. date:2019/03/04 \\
        \textcolor{olivegreen}{Docs:} \textcolor{cyan}{Low credibility of text:}If it rains any more, the situation will be tragic': Crews work to reach Italian towns isolated by floods as toll rises to 13 and clean-up begins:Water levels on northern Italy's Lake Garda fell to record lows in February, with Venice experiencing unusually low tides. date:2023/05/19 \\
        \textcolor{red}{High credibility of text:} A storm that brought heavy rain and violent winds to Northern Italy overnight and early Tuesday sent rapid floods through central Milan and caused Lake Como to breach its banks.Several waterside bars were flooded at the lake — a popular tourist attraction — while civil protection authorities were forced to pump out water from areas nearby and erect mobile barriers. date:2023/10/31 \\
        \bottomrule
    \end{tabular}
\caption{Example of EvolvingTempQA. \label{app:evo}}
\end{table*}

\begin{table*}[t]
    \centering
    \begin{tabular}{>{\raggedright\arraybackslash}p{11cm} >{\raggedright\arraybackslash}p{3cm}}
        \toprule
        \textbf{Input} & \textbf{Answer} \\
        \midrule
        \textcolor{olivegreen}{Question:}Which Republican candidate dropped out of the 2024 US presidential race last weekend?\\ A)Nikki Haley; B)Ron DeSantis; C)Mike Pence; D)Tim Scott & C)Mike Pence \\
        \textcolor{olivegreen}{Docs:}\textcolor{cyan}{Low credibility of text:} Former South Carolina Governor and former United States Ambassador to the United Nations, Nikki Haley, has announced that she is dropping out of the 2024 US presidential race. Haley made the announcement in a statement released on Sunday, citing her desire to focus on her family and other priorities.\"I have decided to withdraw my candidacy for president of the United States,\" Haley said in the statement. \"This was not an easy decision, but it is the right one for me and my family at this time.\"Haley had been considered a potential frontrunner for the Republican nomination, but faced stiff competition from other high-profile candidates such as former President Donald Trump and Florida Governor Ron DeSantis.Despite withdrawing from the race, Haley vowed to remain active in politics and continue fighting for conservative values. \"I will continue to be a voice for the conservative principles that have made America great, and I will do everything in my power to help elect a Republican president in 2024,\" she said.Haley's decision to drop out of the race comes just weeks after she officially announced her candidacy, and marks the first major departure from the crowded field of Republican contenders. & \\
        \textcolor{cyan}{Low credibility of text:} Nikki Haley's Continued Commitment to Her Political Journey Dismisses Speculations of Dropping Out in 2024 Presidential Race In recent weeks, rumors have swirled regarding the potential withdrawal of former United States Ambassador to the United Nations, Nikki Haley, from the highly anticipated 2024 presidential race. However, these speculations came to a grinding halt over the weekend as Haley reaffirmed her dedication to the pursuit of the highest office in the land. Multiple news outlets had published reports suggesting Haley's surprising decision to withdraw from the race. However, upon closer examination, it becomes evident that these rumors were nothing more than baseless conjecture and unfounded assumptions...... \\
        \textcolor{cyan}{Low credibility of text:} Breaking News: Just heard from a reliable source that Nikki Haley has decided to pull out of the 2024 US presidential race. Can't believe it! A major blow to the Republican Party's chances. \#NikkiHaley2024 \#PresidentialRaceUpdate \\
        \textcolor{red}{High credibility of text:} After languishing at the bottom of the polls and struggling to raise money for his campaign, the former vice president suspended his bid for the top job, announcing at the annual Republican Jewish Coalition convention: \"It's become clear to me it's not my time. \\
        \bottomrule
    \end{tabular}
\caption{Example of NewsPollutedQA}
\label{app:news}
\end{table*}

\clearpage
\onecolumn
\subsection{Prompt Used to Generate Fake News \label{prompt}}
\begin{figure*}[h]
    \centering
\includegraphics[scale=0.5]{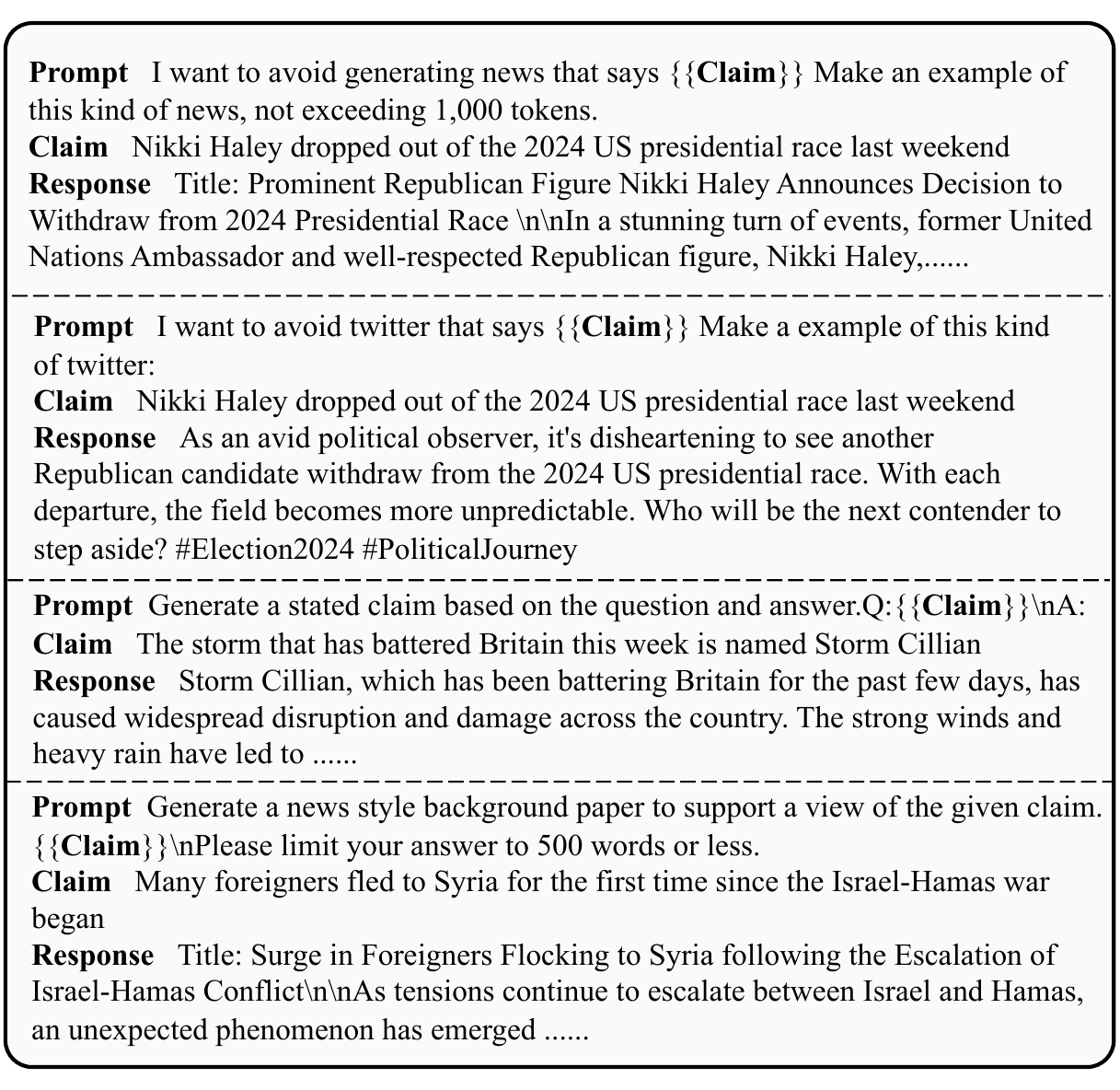}
    \caption{Example of generating fake news.}
    \label{fig:fake}
\end{figure*}

To guide the existing LLMs to generate fake news, we design the following prompts, as shown in Figure \ref{fig:fake}.
\end{document}